\documentclass[10pt,twocolumn,letterpaper]{article}

\usepackage{cvpr}              
\usepackage{makecell}

\definecolor{cvprblue}{rgb}{0.21,0.49,0.74}
\usepackage[pagebackref,breaklinks,colorlinks,allcolors=cvprblue]{hyperref}
\usepackage{dsfont}

\usepackage[table,xcdraw]{xcolor}

\def\paperID{33} 
\def\confName{CVPR}
\def\confYear{2026}

\title{Analyzing Training-Free Corruption Detection for Object Detection Datasets}

\author{
Christian Sieberichs $^{1}$ \quad
Simon Geerkens $^{1}$ \quad
Thomas Waschulzik $^{2}$ \quad\\
Viswanathan Ramesh $^{3}$ \quad
Alexander Braun $^{1}$\\[0.5em]
$^{1}$University of Applied Sciences Düsseldorf \quad
$^{2}$Siemens Mobility GmbH \quad \\
$^{3}$Goethe University Frankfurt\\[0.5em]
}

\begin{document}
\maketitle
\begin{abstract}
Annotation errors are widespread in computer vision datasets and can significantly degrade the performance of systems trained on them, particularly in complex tasks such as object detection. Several approaches exist to identify annotation errors, including training-free feature-space methods which provide a fast and interpretable way to analyze annotations. However, the behavior on object detection annotations, which include semantic and spatial information, remains largely unexplored.

In this work we analyze the applicability of feature-space-based approaches for detecting annotation errors in object detection datasets. By adapting an existing feature-space method, we show that such approaches reliably expose semantic mislabel, while positional errors remain difficult to detect. We evaluate this behavior across multiple pretrained embedding models, synthetic noise types (symmetric, asymmetric, and positional), and real-world annotation errors using VOC2012 and KITTI.

All code and real-world corruptions are publicly available at the following repository: https://github.com/ ChristianSieberichs/BoundingBox\_corruption\_detection 
\end{abstract}

\section{Introduction}
\label{sec:introduction}

High-quality annotations are essential for training reliable machine learning systems. Although self-supervised pretraining has reduced the dependence on large labeled datasets, accurate annotations remain critical for task definitions and fine-tuning computer vision models \cite{GONG2023107268, emam2021statedatacomputervision, DBLP_journals_corr_abs_2002_05709}. Yet human annotations are inherently error-prone. Task complexity, annotator expertise, and ambiguous image content routinely introduce inconsistencies, even in seemingly simple classification datasets, where error rates can reach up to 10\% \cite{northcutt2021pervasivelabelerrorstest}. Such inaccuracies degrade model performance by injecting noise into the training signal \cite{xia2021robust, liu2021understandinginstancelevellabelnoise, 9857348}. As datasets continue to grow in scale and complexity, systematic dataset auditing and annotation quality analysis have become increasingly important.

A growing body of work therefore focuses on automated dataset auditing and corruption detection. Most existing approaches train ML-models on noisy data and identify suspicious instances through predictions, loss values, or gradient behavior \cite{northcutt2022confidentlearningestimatinguncertainty, tkachenko2023objectlabautomateddiagnosismislabeled, chachula2023combatingnoisylabelsobject}. While effective, this approach is computationally demanding and risks inheriting biases from the noisy supervision \cite{liu2021understandinginstancelevellabelnoise}. An alternative are training-free feature-space approaches that analyze neighborhood relationships in embeddings produced by pretrained models. While such approaches may struggle to capture subtle distinctions between visually similar instances, they are computationally efficient, easy to apply across datasets, fast, and provide an interpretable signal through neighborhood relationships within a feature space. 

However, most training-free corruption detection approaches focus on classification tasks in combination with high synthetic noise rates. This setting does not reflect the complexity of real-world vision tasks, which frequently involve multiple objects per image, spatial annotations, and heterogeneous sources of annotation error \cite{cheng2021learninginstancedependentlabelnoise, huang2024learninginstancedependentnoisylabels, Muller_2019}. Object detection datasets in particular introduce additional challenges, as annotations encode semantic labels and spatial localization for multiple instances within a single image.
As a consequence, the prevalence and structure of annotation errors in object detection datasets remains poorly understood. While studies in image classification estimate label noise rates of up to 38\%, the extent of annotation variation in object detection datasets has not yet been systematically quantified \cite{schubert2023identifyinglabelerrorsobject, tschirschwitz2025labelconvergencedefiningupper}. 

In this work, we analyze feature-space-based corruption detection for object detection datasets. Our experiments cover multiple embedding models, corruption types, and datasets ranging from synthetic noise to real-world annotation inconsistencies.
Our main contributions are:
\begin{itemize}
\item We present a systematic analysis of annotation noise in object detection, evaluating label and positional corruptions under controlled synthetic noise as well as real-world dataset conditions
\item We adapt training-free feature-similarity corruption detection to instance-level object detection annotations
\item We identify key strengths and limitations of feature-space corruption detection, including its sensitivity to embedding choice, class imbalance, and spatial perturbations.
\item We provide curated splits of VOC2012 and KITTI along with code and inspection tools to support further research on dataset auditing and annotation quality.
\end{itemize}

\section{Related Work}
\label{sec:relatedwork}

This section reviews dataset analysis approaches for the identification of erroneous annotations within classification and object detection datasets

\subsection{Training-Dependent Methods for Error Detection}
Training-dependent approaches represent the dominant class of methods for the identification of annotation errors. These techniques train a ML-model on potentially erroneous datasets and subsequently use its predictions, confidence scores, loss trajectories, or gradient behavior to flag suspicious instances. While effective, such approaches inherit two structural limitations. They require substantial computational resources, and their performance can be biased by the errors present in the training data.

Confident Learning (CL) is one of the most widely used methods for error detection in classifications \cite{northcutt2022confidentlearningestimatinguncertainty}. Using model predictions, CL estimates the transition matrix and flags potentially mislabeled annotations. CL has been used for the evaluation of ten benchmark datasets, detecting error rates ranging from 0.15\% (MNIST) to 10.12\% (QuickDraw) \cite{northcutt2021pervasivelabelerrorstest}. CL requires a trained model with reliable probability estimates and depends heavily on the quality of the learned decision boundary.

The cleanlab environment extends the applicability of CL beyond classification \cite{cleanlab_docs, cleanlab_object_detection, cleanlab_research}. ObjectLab \cite{tkachenko2023objectlabautomateddiagnosismislabeled} and CLOD \cite{chachula2023combatingnoisylabelsobject} adapt training-dependent corruption detection to object detection.
Both methods compare predicted bounding boxes with annotated ones to identify potential annotation errors. ObjectLab assigns per-instance quality scores for the error types \textit{overlooked}, \textit{swapped}, and \textit{badly located}, and aggregates them into image-level metrics. CLOD applies CL directly to detection data by matching bounding boxes, assigning a predicted label, and treating them as independent instances which are ranked and pruned.
Although both approaches demonstrate strong empirical performance, they require training detection models, which introduces substantial computational overhead and makes the results sensitive to model bias under noisy supervision.

\subsection{Training-Free Methods for Error Detection}
Training-free approaches detect annotation errors without training a ML-model. These methods rely on the representation of image data in a feature space, which makes them computationally efficient and reduces the risk of propagating biases from noisy supervision.

SimiFeat \cite{zhu2022detectingcorruptedlabelstraining} is a lightweight, training-free approach for the flagging of annotation errors in classification datasets. The method embeds samples into a feature space and evaluates local label agreement within a neighborhood. Based on a quality score, instances whose local feature structure contradicts their label are flagged. A detailed description of the approach is provided in section \ref{subsec:SimiFeat}.

Recent work has also explored the use of large multimodal large language models (MLLMs) for detecting erroneous annotations through reasoning. Methods such as VDC \cite{zhu2024vdcversatiledatacleanser} and AutoVDC \cite{vasa2025autovdcautomatedvisiondata} formulate error detection as a visual question-answering task, using models such as BLIP \cite{dai2023instructblipgeneralpurposevisionlanguagemodels} to validate or refute annotations. While not requiring task-specific training, this approach depends on large, resource-intensive models, manually constructed question–answer catalogs, and computationally expensive.

Feature-similarity-based approaches, such as SimiFeat, offer a fast, lightweight, and interpretable alternative for large-scale dataset inspection. However, existing feature-space methods have been developed exclusively for classification datasets. The applicability of feature-space-based corruption detection to object detection, which include semantic labels and spatial information, has not yet been systematically analyzed.

\section{Methods and Experimental Setup}
\label{sec:methods}
This section introduces the feature-space–based analysis pipeline used to study annotation errors in object detection datasets. We also describe the datasets and evaluation protocol employed in our experiments.

\subsection{SimiFeat}
\label{subsec:SimiFeat}
To identify corruptions we rely on a feature-space–based corruption detection pipeline inspired by SimiFeat \cite{zhu2022detectingcorruptedlabelstraining}. 

Let $\mathbf{D}:=(x_n, \tilde{y}_n)_{n\in[N]}$ be a dataset comprising $N$ instances $[N]:=\{1,2,\ldots,N\}$. $x_n$ represents a feature vector $x_n\in\mathbf{X}$ and $y_n$ the corresponding, but unknown, ground truth label $y_n\in\mathbf{Y}$. In most practical cases, clean labels are not available and only the noisy dataset $\mathbf{\tilde{D}}$ containing noisy labels $\tilde{y}_n\in\mathbf{\tilde{Y}}$ is given. When the true label $y_n$ is not identical to the noisy label $\tilde{y}_n$, the instance is considered to be corrupted. SimiFeat focuses on a closed-set in which ground truth and noisy label are part of the same set of possible labels $\mathbf{Y}, \mathbf{\tilde{Y}}\in[K]$.

SimiFeat detects corrupted instances through local label homogeneity by applying a suitable feature representation in which “nearby features should belong to the same true class with a high probability” \cite[p.~3]{zhu2022detectingcorruptedlabelstraining} \cite{disalvo2025embeddingworththousandnoisy,cheng2022instancedependentlabelnoiselearningmanifoldregularized}. Instances whose labels deviate from the labels of their nearest neighbors are likely to be corrupted. This approach avoids the need for model training, but it is sensitive to the choice of feature extractor and distance metric.

Corruption detection is formulated as a neighborhood-consistency problem. For each instance, the label distribution among its $k=10$ nearest neighbors in feature space is computed. This neighborhood distribution forms the basis for two variants of the method.

The simpler variant, SimiFeat-V, flags instances whose label differs from the most frequent label among their neighbors. The second SimiFeat-R variant assigns a continuous corruption score based on the cosine similarity between the neighborhood label distribution and the one-hot encoding of the instance’s label:

\vspace{-10pt}
\begin{equation}
\label{equ:score}
Score(\hat{\textbf{y}}_n,\tilde{\textbf{y}}_n=j)=\frac{\hat{\textbf{y}}^T_n\cdot\textbf{e}_j}{\|\hat{\textbf{y}}_n\|_2\|\textbf{e}_j\|_2},n\in [N],j\in[K]
\end{equation}

Here $\hat{y}_n$ denotes the normalized label distribution among the $k$ nearest neighbors and $\textbf{e}_j$ is the one-hot vector corresponding to the noisy label $\tilde{y}_n$. A score close to one indicates local label agreement, whereas lower scores indicate disagreement with the surrounding neighborhood.

To determine class-specific thresholds, SimiFeat-R employs the High-Order-Consensus (HOC) estimator \cite{zhu2021clusterabilityalternativeanchorpoints}. HOC estimates a transition matrix $\mathbf{T}$ which encodes on its main diagonal the probability that a given class label is correct. This is used as the fraction of correct labeled samples per class. The thereby created class-wise thresholds for the corruption scores can account for imbalanced amounts of data but the assumptions of nearest-neighbor agreement, which the HOC builds on, does not need to be satisfied and may affect the accuracy of $\mathbf{T}$.

SimiFeat was evaluated on the classification datasets CIFAR-10, CIFAR-100, and Clothing1M \cite{krizhevsky2009learning,kaneko2019labelnoiserobustgenerativeadversarial} under synthetic and real-world label noise. Synthetic noise was tested with noise rates between 30\% (instance) and 60\% (symmetric). SimiFeat-R consistently outperformed SimiFeat-V, achieving the best results under symmetric noise.
While the approach proved effective for identifying label inconsistencies, it has not yet been studied in the context of object detection, where annotations also include spatial information.

\subsection{Application of SimiFeat on Instance-Level Object Detection}
\label{subsec:application}
To analyze feature-space-based corruption detection in object detection datasets, we apply the SimiFeat pipeline as a representative similarity-based method. Due to its superior performance, we focus on the SimiFeat-R variant.

In object detection, multiple objects may appear within a single image, each defined by a separate bounding box. To apply SimiFeat at the instance level, each bounding box region is cropped from the image. The region crops are processed independently by the feature extractor. This ensures that each feature vector represents a single object instance rather than a mixture of multiple objects.

The original SimiFeat implementation employs a pretrained ResNet-34 as a feature extractor, which is reasonable for classification datasets with limited visual variability. Object detection datasets, however, contain a wider range of object scales, viewpoints, and contextual variations. To analyze how feature representations influence corruption detection, we evaluate several embedding models of different architectures and sizes: CLIP-ViT-B/32, CLIP-ViT-L/14, DINO-ViT-S/16, and DINOv2-ViT-L/14 \cite{radford2021learningtransferablevisualmodels, github_clip,caron2021emergingpropertiesselfsupervisedvision, oquab2024dinov2learningrobustvisual, github_dino}. These models are trained on large-scale datasets using self-supervised or multimodal objectives and are not restricted to predefined label sets, enabling robust feature representations across diverse visual domains. After feature extraction, the original scoring pipeline is retained. This includes the score calculation \ref{equ:score} as well as the application of cosine similarities for distance calculation and the usage of the $k=10$ nearest neighbors.

\begin{figure}[t]
  \centering

  \begin{minipage}[b]{0.49\columnwidth}
    \centering
    \includegraphics[width=\linewidth]{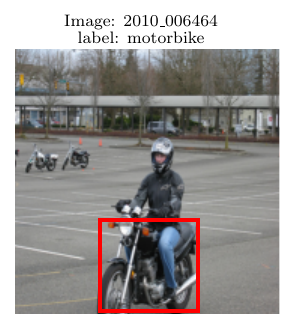}
  \end{minipage}
  \hfill
  \begin{minipage}[b]{0.49\columnwidth}
    \centering
    \includegraphics[width=\linewidth]{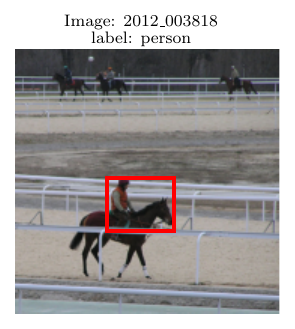}
  \end{minipage}

  \vspace{0em}

  \begin{minipage}[b]{0.49\columnwidth}
    \centering
    \includegraphics[width=\linewidth]{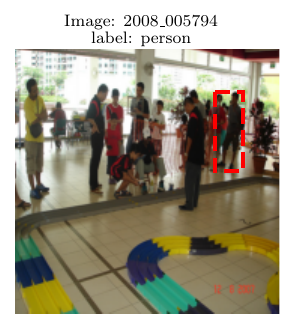}
  \end{minipage}
  \hfill
  \begin{minipage}[b]{0.49\columnwidth}
    \centering
    \includegraphics[width=\linewidth]{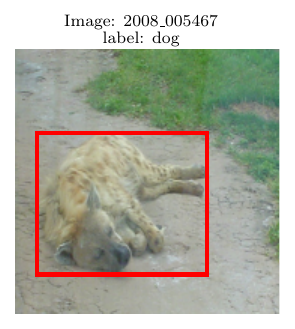}
  \end{minipage}

  \caption{Examples of annotation inconsistencies in the VOC2012 dataset: (1) mislabel, (2) badly located bounding box, (3) overlooked object without annotation, and (4) other corruption (e.g., unseen class labeled as a known category).}
  \label{fig:corruption_example}
  \vspace{-10pt}
\end{figure}

Application of feature-space-based approaches to object detection introduces a wider range of potential corruptions. Following the taxonomy proposed in ObjectLab \cite{tkachenko2023objectlabautomateddiagnosismislabeled}, we categorize corruptions into four categories - see \cref{fig:corruption_example}:

\begin{itemize}
    \item Mislabel (Swapped): the assigned label is inconsistent with the depicted object
    \item Badly located: the bounding box does not accurately enclose the object
    \item Overlooked: an object is present in the image but no annotation is provided
    \item Other: corruptions which are not covered by the other categories
\end{itemize}


Feature-space approaches naturally target mislabels by identifying semantic inconsistencies within local neighborhoods. Positional corruptions introduce a distinct challenge. A badly located bounding box becomes detectable only if the spatial misalignment alters the feature-space position sufficiently to change the neighborhood relationship. This depends on the robustness of the feature extraction process to spatial variations. Models that enforce strong semantic consistency may overlook small localization deviations. Models with stronger spatial sensitivity on the other hand may expose positional errors but potentially at the cost of reduced label stability. Overlooked objects, lacking bounding boxes entirely, remain inherently undetectable for methods relying on existing annotations.

To analyze the types of corruptions detected by the pipeline, all flagged bounding boxes are manually inspected and assigned to one of five categories: mislabel, badly located, truncated or covered, correct, and other. While the categorization involves some subjective judgment, it enables a systematic comparison of corruption types and model behavior across datasets.


\subsection{Experimental Setup}
Our evaluation consists of three stages of increasing complexity. (1) We first reproduce the original SimiFeat results on the CIFAR-10 dataset to compare the original ResNet-34 with modern embedding models. (2) We extend the analysis to object detection using a quality-controlled version of the Pascal VOC2012 dataset with synthetic label and positional corruptions. (3) Finally, we assess the method on naturally occurring corruptions in the KITTI dataset.

We distinguish between label noise and positional noise. For label noise we consider two forms of synthetic corruption commonly used in prior work:
\begin{itemize}
    \item Symmetric noise: an x\% fraction of labels is replaced at random with different labels
    \item Asymmetric noise: an x\% fraction of labels is reassigned to the next label class, following the mapping of the original SimiFeat pipeline
\end{itemize}
Positional noise is introduced by independently perturbing the top-left and bottom-right corners of bounding boxes until a predefined IoU with the original box is reached. This decouples the noise rate (fraction of perturbed boxes) from the noise strength (IoU). This corruption approach creates a wider range of positional corruptions than the mere shift of the original bounding box.
Corruption detection is evaluated as a binary decision problem using the F1-score:

\begin{equation}
    F1=\frac{2}{(Precision^{-1}+Recall^{-1})}
\end{equation}

with:

\begin{equation}
    Precision=\frac{\sum_{n\in[N]}\mathds{1}(v_n=1, \tilde{y}_n\ne y_n)}{\sum_{n\in[N]}\mathds{1}(v_n=1)}
\end{equation}

\begin{equation}
    Recall=\frac{\sum_{n\in[N]}\mathds{1}(v_n=1, \tilde{y}_n\ne y_n)}{\sum_{n\in[N]}\mathds{1}(\tilde{y}_n\ne y_n)}
\end{equation}

Here $\mathds{1}$ is the indicator function. $v_n=1$ indicates that $\tilde{y}_n$ is flagged as corrupted, while $tilde{y}_n\ne y_n $ denotes instances whose annotations are corrupted.

The following sections provide an overview of each dataset used in our experiments.

\textbf{CIFAR10}: 
CIFAR10 \cite{CIFAR_dataset, krizhevsky2009learning} is used to reproduce the results reported in \cite{zhu2022detectingcorruptedlabelstraining} and to create a baseline to compare the tested models with each other. The dataset consists of 50,000 training images and 10,000 test images uniformly distributed across 10 classes.

\textbf{Pascal VOC2012}: 
To evaluate SimiFeat in the object detection setting, we use a quality-controlled version of the Pascal VOC2012 dataset \cite{pascal-voc-2012, hasty}. This dataset removes known annotation errors, providing a clean baseline on which synthetic corruptions can be added in a controlled manner.
The cleaned version of the dataset contains 43,294 bounding boxes in 17,119 images. The dataset corrects a total of 3,215 overlooked, 53 badly located, and 13 mislabeled bounding boxes from the original release. It should be noted that the bounding boxes were counted as badly located if a corrected bounding box was moved by at least 5 pixels in any direction.
The dataset consists of 20 classes and is very unbalanced. Least common are the classes \textit{Bus} (700) and \textit{Train} (706) while the class \textit{Person} is by far the most common (19,330). 

\textbf{KITTI}:
To analyze corruption detection under real noise, we use the \textit{2D object detection evaluation} benchmark of the KITTI dataset \cite{Geiger2012CVPR}. KITTI consists of 7,481 images with 51,853 bounding boxes distributed across nine classes.
Due to its age and evolving annotation conventions, KITTI exhibits several known sources of annotation noise. Bounding boxes may reflect where an object is expected to be rather than its exact visible extent, and occlusion and truncation are handled inconsistently across images. These characteristics make KITTI suitable for studying corruption detection under realistic annotation conditions.

\section{Experimental Analysis} 
We analyze the corruption detection pipeline across datasets of increasing complexity. We begin with CIFAR-10 to reproduce the original SimiFeat results and establish a baseline for comparing feature extractors. Next, we evaluate the method on a cleaned version of the VOC2012 dataset under controlled synthetic label and positional corruptions. Finally, we apply the pipeline to KITTI to analyze naturally occurring annotation inconsistencies.

\subsection{Recreation of SimiFeat on CIFAR}
To validate our implementation and establish a controlled baseline for comparing the added feature extractors, we reproduce the results of the original SimiFeat study. While the original evaluation focuses on noise rates up to 60\%, we examine a broader and more realistic range following \cite{northcutt2021pervasivelabelerrorstest}.
Symmetric and asymmetric label noise are applied following the original SimiFeat implementation \cite{SimiFeat_code}. We evaluate noise rates between 1\% and 40\% and test the method using the original ResNet-34 feature extractor as well as two CLIP and two DINO embedding models.

\begin{figure}[tb]
    \centering
    \includegraphics[width=0.475\textwidth]{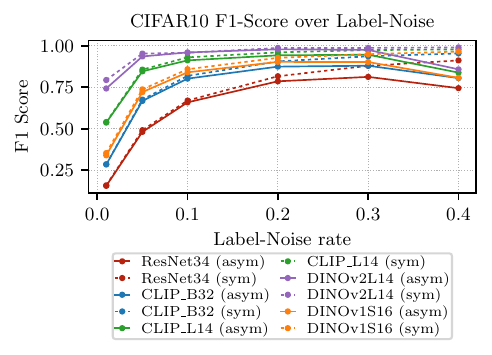}
    \vspace{-20pt}
    \caption{F1-score of different feature extractors on the CIFAR-10 dataset under increasing synthetic label noise. Symmetric and asymmetric noise are applied at different corruption rates. Performance improves with higher noise levels as the proportion of true positives increases.}
    \label{fig:CIFAR_label}
\end{figure}

The results are shown in \cref{fig:CIFAR_label}. For ResNet-34 at an asymmetric noise rate of 30\%, our reproduction closely matches the performance reported in the original publication, achieving an F1-score of 0.81 compared to the reported 0.8358. Although we did not replicate the 60\% symmetric-noise experiment reported in the original work (F1 = 95.56) we got a rising F1 score of 0.91 at a 40\% noise rate, thereby validating the correctness of our implementation.

All evaluated models exhibit the similar behavior under both noise types. At low noise rates, the pipeline performs poorly due to a large number of false positives, which arises because the HOC is overestimating the number of existing corruptions. As the noise rate increases, the proportion of true positives increases, resulting in higher performance. For symmetric noise, the F1-score increases steadily with the rising noise rate. In contrast, for asymmetric noise, performance begins to deteriorate at higher noise rates due to multiple identical corruptions within the same neighborhood. The differences between the evaluated models arise from varying false-positive rates. Higher false-positive rates translate to larger performance drops at low noise levels, at which clean samples dominate.

These findings confirm that SimiFeat performs reliably under high synthetic noise rates, consistent with the original study. However, at low noise rates the method systematically overestimates the number of corrupted samples due to false positives. This behavior occurs across all evaluated embedding models, although the severity varies depending on the quality of the feature representation.

\subsection{Controlled Corruption Analysis on Pascal VOC2012}
To evaluate SimiFeat on object detection data, we use a cleaned version of the Pascal dataset \cite{hasty}. Before introducing artificial noise, we apply our pipeline to the cleaned VOC2012 dataset to quantify the baseline false-positive rate and identify any remaining annotation errors if present. The set of detected bounding boxes is manually inspected and categorized into one of five groups based on Sec. \ref{subsec:application}. Correct annotations are presented more detailed by including truncated or covered instances separately, while overlooked annotations are excluded, as the method operates only on annotated bounding boxes.

\begin{itemize}
    \item Mislabel: the assigned class label was incorrect while the box position was plausible
    \item Badly located: the bounding box failed to tightly enclose the object
    \item Truncated or covered (but correct): the object was only partially visible, yet the annotation remained valid
    \item Correct: the annotation is correct, and the object is clearly identifiable
    \item Others: all remaining cases that did not fit the previous categories, including empty boxes and unlisted objects
\end{itemize}

Despite quality control, the dataset still contains residual annotation errors \ref{tab:manual_VOC_nonoise}. The number of flagged samples ranges from 1794 (DINOv2-ViT/L-14) to 4671 (ResNet-34), corresponding to 3.4\%–8.8\% of all bounding boxes. Although the dataset is nominally clean, 63 genuine corruptions are identified. All identified corruptions are removed before introducing synthetic noise, reducing the datasets size to 43,231 images. Most remaining detections correspond to truncated or occluded objects.

\begin{table}[tb]
    \centering
    \caption{Manual inspection of corruptions in the cleaned VOC2012 dataset without synthetic corruption. The row \textit{Overall} highlights the sum of truely existing corruptions within the dataset.}
    \resizebox{0.475\textwidth}{!}{%
    \begin{tabular}{|l|l|l|l|l|l|l|}
        \hline
        Model         & Detections & Mislabel & \makecell[l]{Badly\\located} & \makecell[l]{Truncated\\or covered} & Correct & Others \\ \hline
        ResNet34      & 4671       & 20       & 11            & 2688                 & 1935    & 17     \\ \hline
        CLIP-ViT-B/32 & 2696       & 19       & 5             & 1748                 & 910     & 14     \\ \hline
        CLIP-ViT-L/14 & 1954       & 18       & 7             & 1333                 & 581     & 15     \\ \hline
        DINO-ViT s16  & 2440       & 16       & 9             & 1487                 & 911     & 17     \\ \hline
        DINOv2 ViTL14 & 1794       & 18       & 4             & 1125                 & 630     & 17     \\ \hline
        Overall       & 63         & 24       & 15            & -                    & -       & 24     \\ \hline
        \end{tabular}%
    }
    \label{tab:manual_VOC_nonoise}
\end{table}

\begin{figure}[bt]
    \centering
    \includegraphics[width=0.475\textwidth]{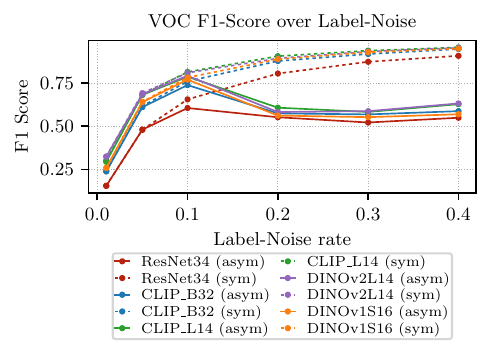}
    \vspace{-20pt}
    \caption{F1-score of tested feature extractors on the VOC2012 dataset under synthetic symmetric and asymmetric label noise. Detection performance increases with higher corruption rates, while asymmetric noise leads to degraded performance due to class imbalance.}
    \label{fig:VOC_symmetricandasymmtric}
\end{figure}

To evaluate corruption detection under controlled conditions, synthetic corruptions are introduced on the cleaned VOC2012 dataset. Label noise follows the CIFAR configuration using symmetric and asymmetric noise rates of 1\%, 5\%, 10\%, 20\%, 30\%, and 40\%.
For positional corruption we vary two parameters: the corruption rate (fraction of perturbed boxes) and the noise strength measured by the IoU between the original and corrupted bounding box. In one experiment the corruption rate follows the label-noise settings with a fixed IoU of 0.4. In a second experiment the corruption rate is fixed at 10\% while the IoU varies across levels of 0.2, 0.4, 0.6 and 0.8. The resulting F1-scores are shown in \cref{fig:VOC_symmetricandasymmtric}.

Across all models, SimiFeat remains computationally efficient. The complete processing pipeline requires between 307s (ResNet-34) and 959s (DINOv2-ViT-L/14), including feature extraction and similarity computation.

Under symmetric label noise, the behavior follows the CIFAR results. The performance is low at small noise rates but increases steadily as the fraction of corrupted annotations grows. At a noise rate of 40\%, all embedding models reach an F1-score of approximately 0.95. ResNet-34 performs consistently worse than the embedding models, highlighting the importance of strong feature representations.

In contrast, asymmetric label noise produces a different pattern. Performance increases until a noise rate of 10\%, after which the F1-scores drop and stabilize near an F1 score of 0.6 across all models. This behavior is cause by the strong class imbalance in the VOC2012 dataset. The class \textit{Person} alone accounts for 19,322 of the 43,231 annotated objects. In the asymmetric configuration, each label is replaced by the next class in a fixed class order. As a consequence, a large number of \textit{Person} instances are relabeled in this context as \textit{Potted plant}, while only a small number of \textit{Potted plant} samples exist to balance this transition. The effective noise therefore becomes highly uneven across classes, reaching extremely high levels for categories with few data points.
This imbalance violates the assumption of the HOC estimator, which infers the expected number of corrupted samples per class through a transition matrix fitted to observed neighborhood relationships. HOC incorrectly estimates similar corruption rates for \textit{Potted plant} and \textit{Person}. As a result HOC overestimates corruption within the dominant class, producing a large number of false positives when the amount of corruptions dominates the true class label. This behavior highlights an important limitation in long-tailed datasets.

\begin{figure}[bt]
    \vspace{-5pt}
    \centering
    \includegraphics[width=0.475\textwidth]{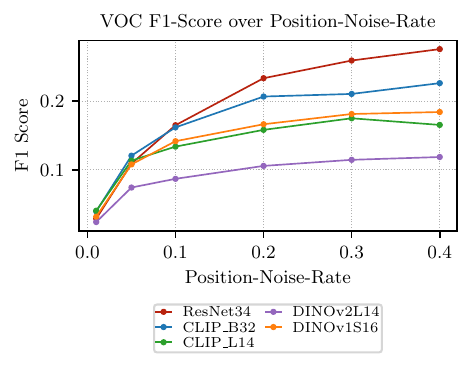}
    \vspace{-24pt}
    \caption{Detection performance under positional noise in VOC2012 and fixed noise strength of 0.4. Performance improves with higher corruption rates but remains substantially lower than for label noise.}
    \label{fig:VOC_positional_noiserate}
    \vspace{-10pt}
\end{figure}

\begin{figure}[bt]
    \centering
    \includegraphics[width=0.475\textwidth]{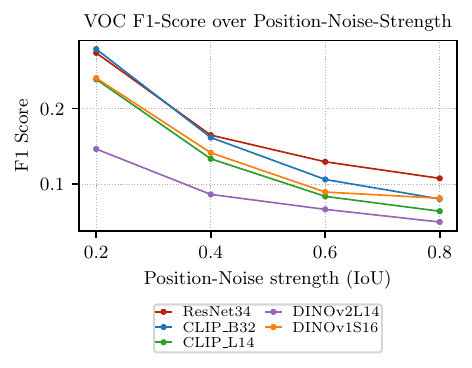}
    \vspace{-24pt}
    \caption{Detection performance under varying positional noise strength in VOC2012 and fixed noise rate of 10\%. Lower IoU values correspond to stronger spatial distortions, which increase detection performance.}
    \label{fig:VOC_positional_noisestrength}
\end{figure}

Positional noise shows a different behavior. Although detection performance increases with higher corruption rates as presented in (\cref{fig:VOC_positional_noiserate}), the absolute performance remains substantially lower than for label noise. Even at a corruption rate of 40\%, many positional corruptions remain undetected, indicating limited sensitivity of feature-space representations to spatial inconsistencies.
The introduction of varying noise strength over a positional noise rate of 10\% in \cref{fig:VOC_positional_noisestrength} shows that the detection performance increases with stronger spatial distortions (smaller IoU).

Interestingly, the ranking of feature extractors differs from the label-noise experiments. While the embedding models outperform ResNet-34 considerably in detecting label inconsistencies, ResNet-34 achieves a good performance for positional corruption detection, while DINOv2-ViT-L/14 performs worst. This inversion suggests that the embeddings capture different aspects of visual similarity. Architectures optimized for strong semantic clustering seem to be effective at identifying label inconsistencies but exhibit limited sensitivity to moderate spatial distortions.

The experiments on VOC2012 demonstrate that feature-space-based corruption detection effectively identifies semantic label inconsistencies but remains less sensitive to positional perturbations. The results also reveal that the choice of embedding model strongly influences the type of corruption that can be detected. Architectures that enforce strong semantic consistency perform well for label noise but struggle to capture positional deviations in bounding box annotations. These findings highlight both the potential and the limitation of feature-space-based approaches for auditing object detection datasets.

\subsection{Analysis on KITTI}
To validate whether the observations obtained under synthetic corruptions also hold for real-world data, we apply the corruption detection pipeline on the KITTI dataset \cite{Geiger2012CVPR}. Unlike the quality controlled version of the VOC2012, KITTI contains naturally occurring corruptions whose frequency and distribution are unknown. To the best of our knowledge, no systematic verification of label quality has been conducted for the dataset to date.

All bounding boxes flagged as suspicious are manually inspected and categorized using the same annotation scheme as in the VOC experiments. Instances that were never flagged remain unverified, resulting in the actual number of corruptions in the entire dataset not being determined.

KITTI contains 51,853 bounding boxes, slightly more than VOC2012 (43,294), and is likewise characterized by class imbalance. As in the previous experiments, the number of detections varies considerably between the models \cref{tab:KITTI_manual_inspection}. ResNet-34 flags 5062 data points, while DINOv2-ViT-L/14 flags 2491. A notable difference appears between the CLIP and DINO embeddings. While the fraction of corrupted instances among flagged samples is comparable, CLIP highlights substantially more instances.

Manual inspection of the flagged instances reveal 1722 unique corruptions, consisting of 1195 mislabel, 197 badly located, and 330 other corruptions \cref{tab:KITTI_manual_inspection}. Many of the cases classified as “other” correspond to bounding boxes without visible objects, often caused by complete occlusion.


Importantly, the models collectively identify 1195 unique mislabeled instances, whereas the best individual model (CLIP-ViT-B/32) detects only 655. This indicates that different feature representations capture complementary inconsistencies in the dataset and that a significant fraction of potential annotation errors remains undiscovered by any single embedding model.

\begin{table}[tb]
    \centering
    \caption{Manual inspection of detected corruptions in the KITTI dataset.}
    \resizebox{0.475\textwidth}{!}{%
    \begin{tabular}{|l|l|l|l|l|l|l|}
        \hline
        Model         & Detections & Mislabel & \makecell[l]{Badly\\located} & \makecell[l]{Truncated\\or covered} & Correct & Others \\ \hline
        ResNet34      & 5062       & 607      & 144           & 1603                 & 2475    & 233    \\ \hline
        CLIP-ViT-B/32 & 4272       & 655      & 68            & 1308                 & 2026    & 215    \\ \hline
        CLIP-ViT-L/14 & 3637       & 636      & 53            & 1215                 & 1513    & 220    \\ \hline
        DINO-ViT-S/16 & 2722       & 451      & 45            & 861                  & 1203    & 162    \\ \hline
        DINOv2-ViT-L/14 & 2491       & 407      & 39            & 820                  & 1022    & 203    \\ \hline
        Overall       & 1722       & 1195     & 197           & -                    & -       & 330    \\ \hline
    \end{tabular}%
    }
    \label{tab:KITTI_manual_inspection}
\end{table}
\begin{table}[tb]
    \centering
    \caption{Manual inspection of detected corruptions in the KITTI dataset for CLIP-ViT-B/32, grouped by annotation.}
    \resizebox{0.475\textwidth}{!}{%
    \begin{tabular}{|l|l|l|l|l|l|l|}
        \hline
        CLIP-ViT-B/32   & Mislabel & \makecell[l]{Badly\\located} & \makecell[l]{Truncated\\or covered} & Correct & Others & Sum  \\ \hline
        Car             & 10       & 14            & 281                  & 221     & 108    & 634  \\ \hline
        Truck           & 9        & 3             & 54                   & 30      & 5      & 101  \\ \hline
        Van             & 34       & 5             & 219                  & 197     & 19     & 474  \\ \hline
        Don’t Care      & 589      & 5             & 461                  & 1099    & 5      & 2159 \\ \hline
        Misc            & 1        & 0             & 94                   & 94      & 8      & 197  \\ \hline
        Pedestrian      & 4        & 23            & 53                   & 136     & 18     & 234  \\ \hline
        Person\_sitting & 5        & 6             & 58                   & 46      & 15     & 130  \\ \hline
        Tram            & 0        & 0             & 20                   & 19      & 1      & 40   \\ \hline
        Cyclist         & 3        & 12            & 68                   & 184     & 36     & 303  \\ \hline
        Sum             & 655      & 68            & 1308                 & 2026    & 215    & 4272 \\ \hline
    \end{tabular}%
    }
    \label{tab:KITTI_manual_perclass}
\end{table}
A per-class analysis of the detected corruptions, exemplified by CLIP-ViT-B/32 in \cref{tab:KITTI_manual_perclass}, shows that 589 of the 655 detected mislabels originate from the class \textit{Don’t Care}. This category represents “regions with unlabeled or ambiguous objects” \cite{KITTI_homepage}.
These regions most often contain small background objects, but they may also contain larger, subjectively identifiable objects that could plausibly belong to one of the labeled classes. When such objects appear close to semantically similar objects within the feature-space, they are interpreted as inconsistencies and potential mislabel. As a consequence, \textit{Don’t Care} accounts for the majority of all identified mislabel and only 100 unique mislabel are corresponding to other classes. This difficulty is one of the reasons why the collective number of detected corruptions is much larger than the best individual one.

Although a quantitative comparison with VOC2012 is not possible due to the absence of ground-truth annotations, the manual inspection reveals a substantial number of annotation inconsistencies in the KITTI dataset. Across all models, mislabels are detected far more frequently than localization errors. While this may reflect a higher proportion of mislabeled data within KITTI, the imbalance is likely influenced by the limited sensitivity of the feature-space pipeline to positional corruptions, consistent with the observations from the controlled VOC experiments. In addition, the majority of detected mislabels originate from the \textit{Don’t Care} class, highlighting the ambiguity of this label and its interaction with visually similar object classes.

\section{Discussion}
\label{sec:discussion}
This study analyzes annotation corruptions in object detection datasets through feature-space-based approaches to explore how annotation inconsistencies manifest within detection datasets. We apply SimiFeat as a representative training-free pipeline.

The experiments provide insights into annotation quality in object detection dataset. Even the quality-controlled VOC2012 dataset still contained residual corruptions, indicating that dataset cleaning pipelines do not eliminate all corruptions. The analysis of the KITTI dataset further illustrates the diversity of annotation issues present in real-world datasets. Manual inspection revealed a substantial number of mislabeled and false positioned bounding boxes, as well as a large fraction of ambiguous cases corresponding to truncated or occluded objects. This observation illustrates how natural visual variation in real-world datasets can be mistaken for annotation inconsistencies. In addition, a significant portion of detected corruptions originated from the \textit{Don't Care} class. Although this class is intended to mark regions that should be ignored during training, many of these regions contain visually identifiable objects that could plausibly belong to one of the labeled categories. The example highlights that lacking annotation quality and even added corrective classes can introduce ambiguity that is difficult to distinguish from true labeling errors.

The experiments also highlight how structural properties of datasets influence the detectability of corruptions. Strong class imbalance, as present in VOC2012 and KITTI, affects the local neighborhood structure used by feature-space analysis. This illustrates how long-tailed class distributions can distort local neighborhood statistics and influence the manifestation of corruptions within a feature space.

Another finding concerns the relationship between semantic and spatial corruptions. Across all datasets, semantic corruptions were detected substantially more reliably than positional corruptions. Small spatial deviations often did not alter the feature-space representation significant enough to flag the corruption, making bounding-box misalignments difficult to detect through feature similarity.

An additional finding is the systematic decline in F1-score at low noise rates or noise strength. This behavior can also be observed in other corruption detection frameworks as well \cite{northcutt2021pervasivelabelerrorstest}, when false positives constitute a substantial portion of detections under small noise rates. But this effect is often somewhat masked in most related studies due to the choice of testing with high artificial noise rates between 20\% and 70\% \cite{tkachenko2023objectlabautomateddiagnosismislabeled, chachula2023combatingnoisylabelsobject, zhu2024vdcversatiledatacleanser}. It is unknown if this effect is present throughout the entire field of corruption detection. 

From a practical perspective, the computational efficiency of training-free corruption detection makes it a useful tool for large-scale dataset auditing. In our experiments, more than 43,000 bounding boxes in VOC2012 could be processed within minutes, allowing suspicious annotations to be efficiently flagged for manual inspection. Such approaches can therefore support dataset maintenance within real time by guiding targeted review rather than attempting to automatically correct datasets.

Overall, our findings highlight both the usefulness and the limitations of feature-space-based corruption detection for flagging annotation quality in object detection datasets. While the approach provides an efficient mechanism for exploring large datasets and identifying potential corruptions, improving sensitivity to spatial corruptions and robustness under low noise conditions remain important directions for future work.

\section{Conclusion}
\label{sec:conclusion}

This work analyzes annotation noise in object detection datasets using feature-space-based corruption detection. Across controlled and real-world datasets, the experiments reveal that annotation inconsistencies extend beyond label corruptions and include spatial misalignments and ambiguous cases caused by truncation, occlusion, or dataset taxonomy. Even quality-controlled datasets such as VOC2012 contained residual corruptions, illustrating the difficulty of fully eliminating annotation errors in large benchmarks.

Our analysis shows that corruption detectability strongly depends on dataset structure and feature representations. Semantic corruptions are reliably exposed while positional corruptions remain difficult to detect due to the limited spatial sensitivity of many feature extractors. Performance further decreases at low corruption rates, where false positives dominate and reduce the overall F1-score. In addition, class imbalance and the choice of feature representation influence which inconsistencies become visible.

Overall, training-free feature-space analysis provides an efficient mechanism for dataset auditing by prioritizing annotations for manual inspection. Future work will focus on improving sensitivity to spatial corruptions and robustness at low corruption rates.

{
\small
\bibliographystyle{ieeenat_fullname}
\bibliography{sources}

@String(CVPR= {IEEE Conf. Comput. Vis. Pattern Recog.})

@String(CVPRW= {IEEE Conf. Comput. Vis. Pattern Recog. Worksh.})

@String(CVPR  = {CVPR})

@String(CVPRW= {CVPRW})

@Misc{northcutt2021pervasivelabelerrorstest,
  author        = {Curtis G. Northcutt and Anish Athalye and Jonas Mueller},
  title         = {Pervasive Label Errors in Test Sets Destabilize Machine Learning Benchmarks},
  year          = {2021},
  archiveprefix = {arXiv},
  eprint        = {2103.14749},
  primaryclass  = {stat.ML},
  url           = {https://arxiv.org/abs/2103.14749},
}

@Misc{northcutt2022confidentlearningestimatinguncertainty,
  author        = {Curtis G. Northcutt and Lu Jiang and Isaac L. Chuang},
  title         = {Confident Learning: Estimating Uncertainty in Dataset Labels},
  year          = {2022},
  archiveprefix = {arXiv},
  eprint        = {1911.00068},
  primaryclass  = {stat.ML},
  url           = {https://arxiv.org/abs/1911.00068},
}

@Misc{zhu2022detectingcorruptedlabelstraining,
  author        = {Zhaowei Zhu and Zihao Dong and Yang Liu},
  title         = {Detecting Corrupted Labels Without Training a Model to Predict},
  year          = {2022},
  archiveprefix = {arXiv},
  eprint        = {2110.06283},
  primaryclass  = {cs.LG},
  url           = {https://arxiv.org/abs/2110.06283},
}

@Misc{zhu2021clusterabilityalternativeanchorpoints,
  author        = {Zhaowei Zhu and Yiwen Song and Yang Liu},
  title         = {Clusterability as an Alternative to Anchor Points When Learning with Noisy Labels},
  year          = {2021},
  archiveprefix = {arXiv},
  eprint        = {2102.05291},
  primaryclass  = {cs.LG},
  url           = {https://arxiv.org/abs/2102.05291},
}

@Misc{cleanlab_docs,
  author = {Cleanlab Team},
  note   = {Accessed: 2025-03-12},
  title  = {Cleanlab Documentation},
  year   = {2024},
  url    = {https://docs.cleanlab.ai/stable/index.html},
}

@Misc{cleanlab_research,
  author = {Cleanlab Team},
  note   = {Accessed: 2025-03-12},
  title  = {Cleanlab Research},
  year   = {2024},
  url    = {https://cleanlab.ai/research/},
}

@Misc{tkachenko2023objectlabautomateddiagnosismislabeled,
  author        = {Ulyana Tkachenko and Aditya Thyagarajan and Jonas Mueller},
  title         = {ObjectLab: Automated Diagnosis of Mislabeled Images in Object Detection Data},
  year          = {2023},
  archiveprefix = {arXiv},
  eprint        = {2309.00832},
  primaryclass  = {cs.CV},
  url           = {https://arxiv.org/abs/2309.00832},
}

@Misc{cleanlab_object_detection,
  author = {Cleanlab Team},
  note   = {Accessed: 2025-03-12},
  title  = {Cleanlab Tutorial: Object Detection},
  year   = {2024},
  url    = {https://docs.cleanlab.ai/stable/tutorials/object_detection.html},
}

@InProceedings{Geiger2012CVPR,
  author    = {Andreas Geiger and Philip Lenz and Raquel Urtasun},
  booktitle = {Conference on Computer Vision and Pattern Recognition (CVPR)},
  title     = {Are we ready for Autonomous Driving? The KITTI Vision Benchmark Suite},
  year      = {2012},
}

@Misc{pascal-voc-2012,
  author       = {Everingham, M. and Van~Gool, L. and Williams, C. K. I. and Winn, J. and Zisserman, A.},
  howpublished = {http://www.pascal-network.org/challenges/VOC/voc2012/workshop/index.html},
  title        = {The {PASCAL} {V}isual {O}bject {C}lasses {C}hallenge 2012 {(VOC2012)} {R}esults},
}

@Misc{cheng2022instancedependentlabelnoiselearningmanifoldregularized,
  author        = {De Cheng and Tongliang Liu and Yixiong Ning and Nannan Wang and Bo Han and Gang Niu and Xinbo Gao and Masashi Sugiyama},
  title         = {Instance-Dependent Label-Noise Learning with Manifold-Regularized Transition Matrix Estimation},
  year          = {2022},
  archiveprefix = {arXiv},
  eprint        = {2206.02791},
  primaryclass  = {cs.LG},
  url           = {https://arxiv.org/abs/2206.02791},
}

@Misc{tschirschwitz2025labelconvergencedefiningupper,
  author        = {David Tschirschwitz and Volker Rodehorst},
  title         = {Label Convergence: Defining an Upper Performance Bound in Object Recognition through Contradictory Annotations},
  year          = {2025},
  archiveprefix = {arXiv},
  eprint        = {2409.09412},
  primaryclass  = {cs.CV},
  url           = {https://arxiv.org/abs/2409.09412},
}

@Misc{schubert2023identifyinglabelerrorsobject,
  author        = {Marius Schubert and Tobias Riedlinger and Karsten Kahl and Daniel Kröll and Sebastian Schoenen and Siniša Šegvić and Matthias Rottmann},
  title         = {Identifying Label Errors in Object Detection Datasets by Loss Inspection},
  year          = {2023},
  archiveprefix = {arXiv},
  eprint        = {2303.06999},
  primaryclass  = {cs.CV},
  url           = {https://arxiv.org/abs/2303.06999},
}

@Misc{hasty,
  author       = {Hasty.ai},
  howpublished = {\url{https://www.edge-ai-vision.com/2022/08/how-we-cleaned-up-pascal-and} \\ \url{-improved-map-by-13/}},
  title        = {How We Cleaned Up {PASCAL} and Improved {mAP} By 13\%},
  year         = {2022},
}

@Online{KITTI_homepage,
  author  = {Karlsruhe Institute of Technology (KIT)},
  title   = {KITTI Vision Benchmark Suite},
  url     = {https://www.cvlibs.net/datasets/kitti/},
  urldate = {2025-04-03},
}

@Misc{kaneko2019labelnoiserobustgenerativeadversarial,
  author        = {Takuhiro Kaneko and Yoshitaka Ushiku and Tatsuya Harada},
  title         = {Label-Noise Robust Generative Adversarial Networks},
  year          = {2019},
  archiveprefix = {arXiv},
  eprint        = {1811.11165},
  primaryclass  = {cs.CV},
  url           = {https://arxiv.org/abs/1811.11165},
}

@TechReport{krizhevsky2009learning,
  author      = {Krizhevsky, Alex},
  institution = {University of Toronto},
  title       = {Learning multiple layers of features from tiny images},
  year        = {2009},
}

@Article{CIFAR_dataset,
  author   = {Alex Krizhevsky and Vinod Nair and Geoffrey Hinton},
  title    = {CIFAR-10 (Canadian Institute for Advanced Research)},
  url      = {http://www.cs.toronto.edu/~kriz/cifar.html},
}

@Misc{liu2021understandinginstancelevellabelnoise,
  author        = {Yang Liu},
  title         = {Understanding Instance-Level Label Noise: Disparate Impacts and Treatments},
  year          = {2021},
  archiveprefix = {arXiv},
  eprint        = {2102.05336},
  primaryclass  = {cs.LG},
  url           = {https://arxiv.org/abs/2102.05336},
}

@Article{GONG2023107268,
  author   = {Youdi Gong and Guangzhen Liu and Yunzhi Xue and Rui Li and Lingzhong Meng},
  journal  = {Information and Software Technology},
  title    = {A survey on dataset quality in machine learning},
  year     = {2023},
  issn     = {0950-5849},
  pages    = {107268},
  volume   = {162},
  doi      = {https://doi.org/10.1016/j.infsof.2023.107268},
  url      = {https://www.sciencedirect.com/science/article/pii/S0950584923001222},
}

@Misc{disalvo2025embeddingworththousandnoisy,
  author        = {Francesco Di Salvo and Sebastian Doerrich and Ines Rieger and Christian Ledig},
  title         = {An Embedding is Worth a Thousand Noisy Labels},
  year          = {2025},
  archiveprefix = {arXiv},
  eprint        = {2408.14358},
  primaryclass  = {cs.CV},
  url           = {https://arxiv.org/abs/2408.14358},
}

@inproceedings{
xia2021robust,
title={Robust early-learning: Hindering the memorization of noisy labels},
author={Xiaobo Xia and Tongliang Liu and Bo Han and Chen Gong and Nannan Wang and Zongyuan Ge and Yi Chang},
booktitle={International Conference on Learning Representations},
year={2021},
url={https://openreview.net/forum?id=Eql5b1_hTE4}
}

@misc{chachula2023combatingnoisylabelsobject,
      title={Combating noisy labels in object detection datasets}, 
      author={Krystian Chachula and Jakub Lyskawa and Bartlomiej Olber and Piotr Fratczak and Adam Popowicz and Krystian Radlak},
      year={2023},
      eprint={2211.13993},
      archivePrefix={arXiv},
      primaryClass={cs.CV},
      url={https://arxiv.org/abs/2211.13993}, 
}

@misc{cheng2021learninginstancedependentlabelnoise,
      title={Learning with Instance-Dependent Label Noise: A Sample Sieve Approach}, 
      author={Hao Cheng and Zhaowei Zhu and Xingyu Li and Yifei Gong and Xing Sun and Yang Liu},
      year={2021},
      eprint={2010.02347},
      archivePrefix={arXiv},
      primaryClass={cs.LG},
      url={https://arxiv.org/abs/2010.02347}, 
}

@misc{huang2024learninginstancedependentnoisylabels,
      title={Learning with Instance-Dependent Noisy Labels by Anchor Hallucination and Hard Sample Label Correction}, 
      author={Po-Hsuan Huang and Chia-Ching Lin and Chih-Fan Hsu and Ming-Ching Chang and Wei-Chao Chen},
      year={2024},
      eprint={2407.07331},
      archivePrefix={arXiv},
      primaryClass={cs.CV},
      url={https://arxiv.org/abs/2407.07331}, 
}

@INPROCEEDINGS{9857348,
  author={Ma, Jiaxin and Ushiku, Yoshitaka and Sagara, Miori},
  booktitle={2022 IEEE/CVF Conference on Computer Vision and Pattern Recognition Workshops (CVPRW)}, 
  title={The Effect of Improving Annotation Quality on Object Detection Datasets: A Preliminary Study}, 
  year={2022},
  volume={},
  number={},
  pages={4849-4858},
  doi={10.1109/CVPRW56347.2022.00532}}

@inproceedings{Muller_2019,
   title={Identifying Mislabeled Instances in Classification Datasets},
   url={http://dx.doi.org/10.1109/IJCNN.2019.8851920},
   DOI={10.1109/ijcnn.2019.8851920},
   booktitle={2019 International Joint Conference on Neural Networks (IJCNN)},
   publisher={IEEE},
   author={Muller, Nicolas M. and Markert, Karla},
   year={2019},
   month=jul, pages={1–8} }

@Misc{SimiFeat_code,
  author       = {UCSC-REAL},
  howpublished = {\url{https://github.com/UCSC-REAL/SimiFeat}},
  title        = {SimiFeat},
  year         = {2025},
}

@misc{emam2021statedatacomputervision,
      title={On The State of Data In Computer Vision: Human Annotations Remain Indispensable for Developing Deep Learning Models}, 
      author={Zeyad Emam and Andrew Kondrich and Sasha Harrison and Felix Lau and Yushi Wang and Aerin Kim and Elliot Branson},
      year={2021},
      eprint={2108.00114},
      archivePrefix={arXiv},
      primaryClass={cs.CV},
      url={https://arxiv.org/abs/2108.00114}, 
}

@misc{zhu2024vdcversatiledatacleanser,
      title={VDC: Versatile Data Cleanser based on Visual-Linguistic Inconsistency by Multimodal Large Language Models}, 
      author={Zihao Zhu and Mingda Zhang and Shaokui Wei and Bingzhe Wu and Baoyuan Wu},
      year={2024},
      eprint={2309.16211},
      archivePrefix={arXiv},
      primaryClass={cs.CV},
      url={https://arxiv.org/abs/2309.16211}, 
}

@Article{DBLP_journals_corr_abs_2002_05709,
  author     = {Ting Chen and Simon Kornblith and Mohammad Norouzi and Geoffrey E. Hinton},
  journal    = {CoRR},
  title      = {A Simple Framework for Contrastive Learning of Visual Representations},
  year       = {2020},
  volume     = {abs/2002.05709},
  bibsource  = {dblp computer science bibliography, https://dblp.org},
  eprint     = {2002.05709},
  eprinttype = {arXiv},
  url        = {https://arxiv.org/abs/2002.05709},
}

@misc{dai2023instructblipgeneralpurposevisionlanguagemodels,
      title={InstructBLIP: Towards General-purpose Vision-Language Models with Instruction Tuning}, 
      author={Wenliang Dai and Junnan Li and Dongxu Li and Anthony Meng Huat Tiong and Junqi Zhao and Weisheng Wang and Boyang Li and Pascale Fung and Steven Hoi},
      year={2023},
      eprint={2305.06500},
      archivePrefix={arXiv},
      primaryClass={cs.CV},
      url={https://arxiv.org/abs/2305.06500}, 
}

@Misc{radford2021learningtransferablevisualmodels,
  author        = {Alec Radford and Jong Wook Kim and Chris Hallacy and Aditya Ramesh and Gabriel Goh and Sandhini Agarwal and Girish Sastry and Amanda Askell and Pamela Mishkin and Jack Clark and Gretchen Krueger and Ilya Sutskever},
  title         = {Learning Transferable Visual Models From Natural Language Supervision},
  year          = {2021},
  archiveprefix = {arXiv},
  eprint        = {2103.00020},
  primaryclass  = {cs.CV},
  url           = {https://arxiv.org/abs/2103.00020},
}

@misc{github_clip,
  author       = {OpenAI},
  title        = {CLIP: Contrastive Language--Image Pretraining (GitHub Repository)},
  howpublished = {\url{https://github.com/openai/CLIP}},
}

@misc{caron2021emergingpropertiesselfsupervisedvision,
      title={Emerging Properties in Self-Supervised Vision Transformers}, 
      author={Mathilde Caron and Hugo Touvron and Ishan Misra and Hervé Jégou and Julien Mairal and Piotr Bojanowski and Armand Joulin},
      year={2021},
      eprint={2104.14294},
      archivePrefix={arXiv},
      primaryClass={cs.CV},
      url={https://arxiv.org/abs/2104.14294}, 
}

@misc{oquab2024dinov2learningrobustvisual,
      title={DINOv2: Learning Robust Visual Features without Supervision}, 
      author={Maxime Oquab and Timothée Darcet and Théo Moutakanni and Huy Vo and Marc Szafraniec and Vasil Khalidov and Pierre Fernandez and Daniel Haziza and Francisco Massa and Alaaeldin El-Nouby and Mahmoud Assran and Nicolas Ballas and Wojciech Galuba and Russell Howes and Po-Yao Huang and Shang-Wen Li and Ishan Misra and Michael Rabbat and Vasu Sharma and Gabriel Synnaeve and Hu Xu and Hervé Jegou and Julien Mairal and Patrick Labatut and Armand Joulin and Piotr Bojanowski},
      year={2024},
      eprint={2304.07193},
      archivePrefix={arXiv},
      primaryClass={cs.CV},
      url={https://arxiv.org/abs/2304.07193}, 
}

@misc{github_dino,
  author       = {Facebook Research},
  title        = {DINO: Self-Supervised Vision Transformers (GitHub Repository)},
  howpublished = {\url{https://github.com/facebookresearch/dino}},
}

@misc{vasa2025autovdcautomatedvisiondata,
      title={AutoVDC: Automated Vision Data Cleaning Using Vision-Language Models}, 
      author={Santosh Vasa and Aditi Ramadwar and Jnana Rama Krishna Darabattula and Md Zafar Anwar and Stanislaw Antol and Andrei Vatavu and Thomas Monninger and Sihao Ding},
      year={2025},
      eprint={2507.12414},
      archivePrefix={arXiv},
      primaryClass={cs.CV},
      url={https://arxiv.org/abs/2507.12414}, 
}
}


\end{document}